\documentclass[10pt,twocolumn,letterpaper]{article}

\usepackage{cvpr}
\usepackage{times}
\usepackage{epsfig}
\usepackage{graphicx}
\usepackage{amsmath}
\usepackage{amssymb}
\usepackage{enumitem}
\usepackage{multirow}

\usepackage[breaklinks=true,bookmarks=false]{hyperref}

\cvprfinalcopy 

\def\cvprPaperID{****} 

\begin{document}


\title{Improved Techniques for Learning to Dehaze and Beyond: A Collective Study}
\author{Yu Liu\textsuperscript{1}, Guanlong Zhao\textsuperscript{2}, Boyuan Gong\textsuperscript{2}, Yang Li\textsuperscript{1}, Ritu Raj\textsuperscript{2}, Niraj Goel\textsuperscript{2}, Satya Kesav\textsuperscript{2}, \\ Sandeep Gottimukkala\textsuperscript{2}, Zhangyang Wang\textsuperscript{2}, Wenqi Ren\textsuperscript{3}, Dacheng Tao\textsuperscript{4}\\
\textsuperscript{1}Department of Electrical and Computer Engineering, Texas A\&M University\\
\textsuperscript{2}Department of Computer Science and Engineering, Texas A\&M University \\
\textsuperscript{3}Chinese Academy of Sciences \qquad
\textsuperscript{4} 
University of Sydney
%
}

\maketitle

\begin{abstract}
\vspace{-1em}
Here we explore two related but important tasks based on the recently released REalistic Single Image DEhazing (RESIDE) benchmark dataset: (i) single image dehazing as a low-level image restoration problem; and (ii) high-level visual understanding (e.g., object detection) of hazy images. For the first task, we investigated a variety of loss functions and show that perception-driven loss significantly improves dehazing performance. In the second task, we provide multiple solutions including using advanced modules in the dehazing-detection cascade and domain-adaptive object detectors. In both tasks, our proposed solutions significantly improve performance. GitHub repository URL: \url{https://github.com/guanlongzhao/dehaze}. 

\end{abstract}
\vspace{-1em}
\section{Introduction}
\label{Intro}
Images taken in outdoor environments affected by air pollution, dust, mist, and fumes often contain complicated, non-linear, and data-dependent noise, also known as haze. Haze complicates many high-level computer vision tasks such as object detection and recognition. 
Therefore, dehazing has been widely studied in the fields of computational photography and computer vision. Early dehazing approaches often required additional information such as the provision or capture of scene depth by comparing several different images of the same scene~\cite{tan2000enhancement,schechner2001instant,kopf2008deep}. Many approaches have since been proposed to exploit natural image priors and to perform statistical analyses~\cite{he2011single,tang2014investigating,zhu2015fast,berman2016non}. Most recently, dehazing algorithms based on neural networks \cite{cai2016dehazenet,ren2016single,li2017aod} have delivered state-of-the-art performance. For example, AOD-Net \cite{li2017aod} trains an end-to-end system and shows superior performance according to multiple evaluation metrics, improving object detection in the haze using end-to-end training of dehazing and detection modules.
\section{Review and Task Description}

Here we study two haze-related tasks: 1) boosting single image dehazing performance as an image restoration problem; and 2) improving object detection accuracy in the presence of haze. As noted by \cite{wang2016studying,li2017aod,li2017reside}, the second task is related to, but is often unaligned with, the first.

While the first task has been well studied in recent works, we propose that \textbf{the second task is more relevant in practice and deserves greater attention}. Haze does not affect human visual perceptual quality as much as resolution, noise, and blur; indeed, some hazy photos may even have better aesthetics. However, haze in unconstrained outdoor environments could be detrimental to machine vision systems, since most of them only work well for haze-free scenes. Taking autonomous driving as an example, hazy and foggy weather will obscure the vision of on-board cameras and create confusing reflections and glare, creating problems even for state-of-the-art self-driving cars \cite{li2017reside}.

\subsection{Haze Modeling and Dehazing Approaches}
The atmospheric scattering model has been widely used to represent hazy images in haze removal works~\cite{mccartney1976optics,narasimhan2000chromatic,narasimhan2002vision}:

\begin{equation}
\label{haze model definition}
I(x) = J(x)t(x) + A(1-t(x)),
\end{equation}

\noindent where $x$ indexes pixels in the observed hazy image, $I(x)$ is the observed hazy image, and $J(x)$ is the clean image to be recovered. The parameter $A$ denotes the global atmospheric light, and $t(x)$ is the transmission matrix defined as:
\begin{equation}
\label{trasmission matrix}
t(x) = e^{-\beta d(x)},
\end{equation}
\noindent where $\beta$ is the scattering coefficient, and $d(x)$ represents the distance between the object and camera.

Conventional single image dehazing methods commonly exploit natural image priors (for example, the dark channel prior (DCP)~\cite{he2011single,tang2014investigating}, the color attenuation prior~\cite{zhu2015fast}, and the non-local color cluster prior~\cite{berman2016non}) and perform statistical analysis to recover the transmission matrix $t(x)$. More recently, convolutional neural networks(CNNs) have been applied for haze removal after demonstrating success in many other computer vision tasks. Some of the most effective models include the multi-scale CNN (MSCNN) which predicts a coarse-scale holistic transmission map of the entire image and refines it locally~\cite{ren2016single}; DehazeNet, a trainable transmission matrix estimator that recovers the clean image combined with estimated global atmospheric light~\cite{cai2016dehazenet}; and the end-to-end dehazing network, AOD-Net \cite{li2017aod,li2017all}, which takes a hazy image as input and directly generates a clean image output. AOD-Net has also been extended to video \cite{li2017end}.

\subsection{RESIDE Dataset}
We benchmark against the REalistic Single Image DEhazing (RESIDE) dataset~\cite{li2017reside}. RESIDE was the first large-scale dataset for benchmarking single image dehazing algorithms and includes both indoor and outdoor hazy images
\footnote{The RESIDE dataset was updated in March 2018, with some changes made to dataset organization. Our experiments were all conducted on the original RESIDE version, now called \href{https://sites.google.com/view/reside-dehaze-datasets}{RESIDE-v0}.}.
Further, RESIDE contains both synthetic and real-world hazy images, thereby highlighting diverse data sources and image contents. It is divided into five subsets, each serving different training or evaluation purposes. RESIDE contains $110,500$ synthetic indoor hazy images (ITS) and $313,950$ synthetic outdoor hazy images (OTS) in the training set, with an option to split them for validation. The RESIDE test set is uniquely composed of the synthetic objective testing set (SOTS), the annotated real-world task-driven testing set (RTTS), and the hybrid subjective testing set (HSTS) containing $1,000$, $4,332$, and $20$ hazy images, respectively. The three test sets address different evaluation viewpoints including restoration quality (PSNR, SSIM and no-reference metrics), subjective quality (rated by humans), and task-driven utility (using object detection, for example). 

Most notably, RTTS is the only existing public dataset that can be used to evaluate object detection in hazy images, representing mostly real-world traffic and driving scenarios. Each image is annotated with object bounding boxes and categories (person, bicycle, bus, car, or motorbike). $4,807$ unannotated real-world hazy images are also included in the dataset for potential domain adaptation.
 
For Task 1, we used the training and validation sets from ITS + OTS, and the evaluation is based on PSNR and SSIM. For Task 2, we used the RTTS set for testing and evaluated using mean average precision (MAP) scores.

\section{Task 1: Dehazing as Restoration}

Most CNN dehazing models \cite{cai2016dehazenet,ren2016single,li2017aod} refer to the mean-squares error (MSE) or $\ell_2$ norm-based loss functions. However, MSE is well-known to be imperfectly correlated with human perception of image quality \cite{zhang2012comprehensive,zhao2017loss}. Specifically, for dehazing, 
the $\ell_2$ norm implicitly assumes that the degradation is additive white Gaussian noise, which is oversimplified and invalid for haze. 
Conversely, $\ell_2$ treats the impact of noise independently of the local image characteristics such as structural information, luminance and contrast. However, according to \cite{wang2004image}, the sensitivity of the Human Visual System (HVS) to noise depends on the local properties and structure of a vision. 
 
Here we aimed to identify loss functions that better match human perception to train a dehazing neural network. We used AOD-Net~\cite{li2017aod} (originally optimized using MSE loss) as the backbone but replaced its loss function with the following options:

\begin{itemize}[noitemsep,topsep=2pt,parsep=2pt,partopsep=2pt]




\item \textbf{$\ell_1$ loss}: The $\ell_1$ loss for a patch $P$ can be written as:

\begin{equation}
\label{l1 error function}
\mathcal{L}^{\ell_1}(P) = \frac{1}{N}\sum_{p \in P}^{} |x(p)-y(p)|.
\end{equation}
where $N$ is the number of pixels in the patch, $p$ is the index of the pixel, and $x(p)$ and $y(p)$ are the pixel values of the generated image and the ground truth image respectively.
\item \textbf{SSIM loss}: Following \cite{zhao2017loss}, we write the SSIM for pixel $p$ as:
\begin{equation}
\label{SSIM}
\begin{split}
SSIM(p) & = \frac{2\mu_x\mu_y + C_1}{\mu_x^2+\mu_y^2+C_1}\cdot \frac{2\sigma_{xy}+C_2}{\sigma_x^2 + \sigma_y^2+C_2} \\
& =l(p)\cdot cs(p).
\end{split}
\end{equation}
The means and standard deviations are computed using a Gaussian filter with standard deviation $\sigma_G$. The loss function for SSIM can then be defined as:
\begin{equation}
\label{SSIM error function}
\mathcal{L}^{SSIM}(P) = \frac{1}{N}\sum_{p \in P}^{} 1 - SSIM(p).
\end{equation}

\item \textbf{MS-SSIM loss}: 
The choice of $\sigma_G$ would impact the training performance of SSIM. Here we adopt the idea of multi-scale SSIM \cite{zhao2017loss}, where $M$ different values of $\sigma_G$ are pre-chosen and fused:
\begin{equation}
\label{MS-SSIM}
\mathcal{L}^{MS-SSIM}(P) = l_M^\alpha(p) \cdot \prod_{j=1}^{M} cs_j^{\beta_j} (P).
\end{equation}

\item \textbf{MS-SSIM+$\ell_2$ Loss}: using a weighted sum of MS-SSIM and $\ell_2$ as the loss function:

\begin{equation}
\label{MSSSIM-L2}
\mathcal{L}^{MS-SSIM-\ell_2} = \alpha \cdot \mathcal{L}^{MSSSIM} + (1-\alpha)\cdot G_{\sigma_{G}^M} \cdot \mathcal{L}^{\ell_2},
\end{equation}

\noindent a point-wise multiplication between $G_{\sigma_{G}^M}$ and $\mathcal{L}^{\ell_2}$ is added for the $\ell_2$ loss function term, because MS-SSIM propagates the error at pixel $q$ based on its contribution to MS-SSIM of the central pixel $\widetilde q$, as determined by the Gaussian weights.

\item \textbf{MS-SSIM+$\ell_1$ loss}: using a weighted sum of MS-SSIM and $\ell_1$ as the loss function:
\begin{equation}
\label{MSSSIM-L1}
\mathcal{L}^{MSSSIM-\ell_1} = \alpha \cdot \mathcal{L}^{MSSSIM} + (1-\alpha)\cdot G_{\sigma_{G}^M} \cdot \mathcal{L}^{\ell_1},
\end{equation}
\noindent the $\ell_1$ loss is similarly weighted by $G_{\sigma_{G}^M}$.

\end{itemize}

We selected 1,000 images from ITS + OTS as the validation set and the remaining images for training. The initial learning rate and mini-batch size of the systems were set to $0.01$ and $8$, respectively, for all methods. All weights were initialized as Gaussian random variables, unless otherwise specified. We used a momentum of $0.9$ and a weight decay of $0.0001$. We also clipped the $\ell_2$ norm of the gradient to be within [-0.1, 0.1] to stabilize network training. All models were trained on an Nvidia GTX 1070 GPU for around 14 epochs, which empirically led to convergence. For SSIM loss, $\sigma_G$ was set to 5. $C_1$ and $C_2$ in (\ref{SSIM}) were 0.01 and 0.03, respectively. For MS-SSIM losses, the multiple Gaussian filters were constructed by setting $\sigma_G^i=\{0.5, 1, 2, 4, 8\}$. $\alpha$ was set as 0.025 for MS-SSIM+$\ell_1$, and 0.1 for MS-SSIM+$\ell_2$, following \cite{zhao2017loss}. 

As shown in Tables \ref{psnr-res-overall} and \ref{ssim-res-overall}, simply replacing the loss functions resulted in noticeable differences in performance. While the original AOD-Net with MSE loss performed well on indoor images, it was less effective on outdoor images, which are usually the images needing to be dehazed in practice. Of all the options, MS-SSIM-$\ell_2$ achieved both the highest overall PSNR and SSIM results, resulting in 0.88 dB PSNR and 0.182 SSIM improvements over the state-of-the-art AOD-Net. We further fine-tuned the MS-SSIM-$\ell_2$ model, including using a pre-trained AOD-Net as a warm initialization, adopting a smaller learning rate (0.002) and a larger minibatch size (16). Finally, the best achievable PSNR and SSIM were 23.43 dB and 0.8747, respectively. Note that the best SSIM represented a nearly 0.02 improvement over AOD-Net. 

\begin{table}[t]
\begin{center}
\begin{tabular}{|l|c|c||c|}
\hline
\multirow{2}{*}{Models}&\multicolumn{3}{c|}{PSNR}\\ \cline{2-4} 
&Indoor&Outdoor&All\\ \hline
AOD-Net Baseline&\textbf{21.01}&24.08&22.55\\ \hline
$\ell_1$&20.27&25.83&23.05\\ \hline
SSIM&19.64&26.65&23.15\\ \hline
MS-SSIM&19.54&\textbf{26.87}&23.20\\ \hline
MS-SSIM+$\ell_1$&20.16&26.20&23.18\\ \hline
MS-SSIM+$\ell_2$&20.45&26.38&\textbf{23.41}\\ \hline
\hline
MS-SSIM+$\ell_2$ (fine-tuned)&20.68&26.18&\textbf{23.43}\\ \hline
\end{tabular}
\end{center}
\caption{Comparison of PSNR results (dB) for Task 1.}
\label{psnr-res-overall}
\end{table}

\begin{table}[t]
\begin{center}
\begin{tabular}{|l|c|c||c|}
\hline
\multirow{2}{*}{Models}&\multicolumn{3}{c|}{SSIM}\\ \cline{2-4} 
&Indoor&Outdoor&All\\ \hline
AOD-Net Baseline &\textbf{0.8372}&0.8726&0.8549\\ \hline
$\ell_1$&0.8045&0.9111&0.8578\\ \hline
SSIM&0.7940&0.8999&0.8469\\ \hline
MS-SSIM&0.8038&0.8989&0.8513\\ \hline
MS-SSIM+$\ell_1$&0.8138&\textbf{0.9184}&0.8661\\ \hline
MS-SSIM+$\ell_2$&0.8285&0.9177&\textbf{0.8731}\\ \hline
\hline
MS-SSIM+$\ell_2$ (fine-tuned)&0.8229 & \textbf{0.9266} &\textbf{0.8747}\\ \hline
\end{tabular}
\end{center}
\caption{Comparison of SSIM results for Task 1.}
\label{ssim-res-overall}
\end{table}


\section{Task 2: Dehazing for Detection}

\subsection{Solution Set 1: Enhancing Dehazing and/or Detection Modules in the Cascade}  

In \cite{li2017aod}, the authors proposed a cascade of AOD-Net dehazing and Faster-RCNN \cite{ren2015faster} detection modules to detect objects in hazy images. We therefore considered it intuitive to try different combinations of more powerful dehazing/detection modules in the cascade. Note that such a cascade could be subject to further joint optimization, as many previous works \cite{liu2017image,cheng2017robust,li2017aod}. However, \textbf{to be consistent with the results in} \cite{li2017reside}, all detection models used in this section were the original pre-trained versions, \textit{without any re-training or adaptation}.

Our solution set 1 considered several popular dehazing modules including DCP \cite{he2011single}, DehazeNet \cite{cai2016dehazenet}, AOD-Net \cite{li2017aod}, and the recently proposed densely connected pyramid dehazing network (DCPDN) \cite{zhang2018densely}. Since hazy images tend to have lower contrast, we also included a contrast enhancement method called contrast limited Adaptive histogram equalization (CLAHE). Regarding the choice of detection modules, we included Faster R-CNN \cite{ren2015faster}\footnote{We replace the backbone of Faster R-CNN from VGG 16 as used by \cite{li2017reside} with the ResNet101 model \cite{he2016deep} to enhance performance.}, SSD \cite{liu2016ssd}, RetinaNet \cite{lin2017focal}, and Mask-RCNN \cite{he2017mask}. 

The compared pipelines are shown in Table \ref{detection1}. In each pipeline, ``X+Y'' by default means applying Y directly on the output of X in a sequential manner. The most important observation is that simply applying more sophisticated detection modules is unlikely to boost the performance of the dehazing-detection cascade, due to the domain gap between hazy/dehazed and clean images (on which typical detectors are trained). The more sophisticated pre-trained detectors (RetinaNet, Mask-RCNN) may have overfitted the clean image domain, again highlighting the demand of handling domain shifts in real-world detection problems. Moreover, a better dehazing model in terms of restoration performance does not imply better detection results on its pre-processed images (e.g., DPDCN). Further, adding dehazing pre-processing does not always guarantee better detection (e.g, comparing RetinaNet versus AOD-Net + RetinaNet), consistent with the conclusion made in \cite{li2017reside}. In addition, AOD-Net tended to generate smoother results but with lower contrast than the others, potentially compromising detection. Therefore, we created two three-stage cascades as in the last two rows of Table \ref{detection1}, and found that using DCP to process AOD-Net dehazed results (with greater contrast) further marginally improved results. 

\begin{table}[t]
\begin{center}
\begin{tabular}{|l|c|}
\hline
\textbf{Pipelines} & \textbf{mAP} \\ 
 \hline
Faster R-CNN & 0.541 \\ \hline
SSD & 0.556 \\ \hline
RetinaNet & 0.531 \\ \hline
Mask-RCNN & 0.457 \\ \hline
\hline
DehazeNet + Faster R-CNN & 0.557 \\ \hline
AOD-Net + Faster R-CNN & \textcolor{blue}{0.563} \\ \hline
DCP + Faster R-CNN & \textcolor{green}{0.567} \\ \hline
DehazeNet + SSD & 0.554 \\ \hline
AOD-Net + SSD & 0.553\\ \hline
DCP + SSD & 0.557 \\ \hline
AOD-Net + RetinaNet & 0.419 \\ \hline
DPDCN + RetinaNet & 0.543 \\ \hline
DPDCN + Mask-RCNN & 0.477 \\ \hline
\hline
AOD-Net + DCP + Faster R-CNN & \textcolor{red}{0.568} \\ \hline
CLACHE + DCP + Mask-RCNN & 0.551 \\ \hline
\end{tabular}
\end{center}
\caption{Solution set 1 mAP results on RTTS. Top 3 results are colored in red, green, and blue, respectively.}
\label{detection1}
\end{table}

\subsection{Solution Set 2: Domain-Adaptive Mask-RCNN} 

Motivated by the observations made on solution set 1, we next aimed to more explicitly tackle the domain gap between hazy/dehazed images and clean images for object detection. Inspired by the recently proposed domain adaptive Faster-RCNN \cite{chen2018domain}, we applied a similar approach to design a domain-adaptive mask-RCNN (DMask-RCNN). 





In the model shown in Figure \ref{fig:DMask}, the primary goal of DMask-RCNN is to mask the features generated by feature extraction network to be as domain invariant as possible, between the source domain (clean input images) and the target domain (hazy images). Specifically, DMask-RCNN places a domain-adaptive component branch after the base feature extraction convolution layers of Mask-RCNN. The loss of the domain classifier is a binary cross entropy loss:

\begin{figure}[h]
\centering
\includegraphics[width=3.5in,height=1.5in]{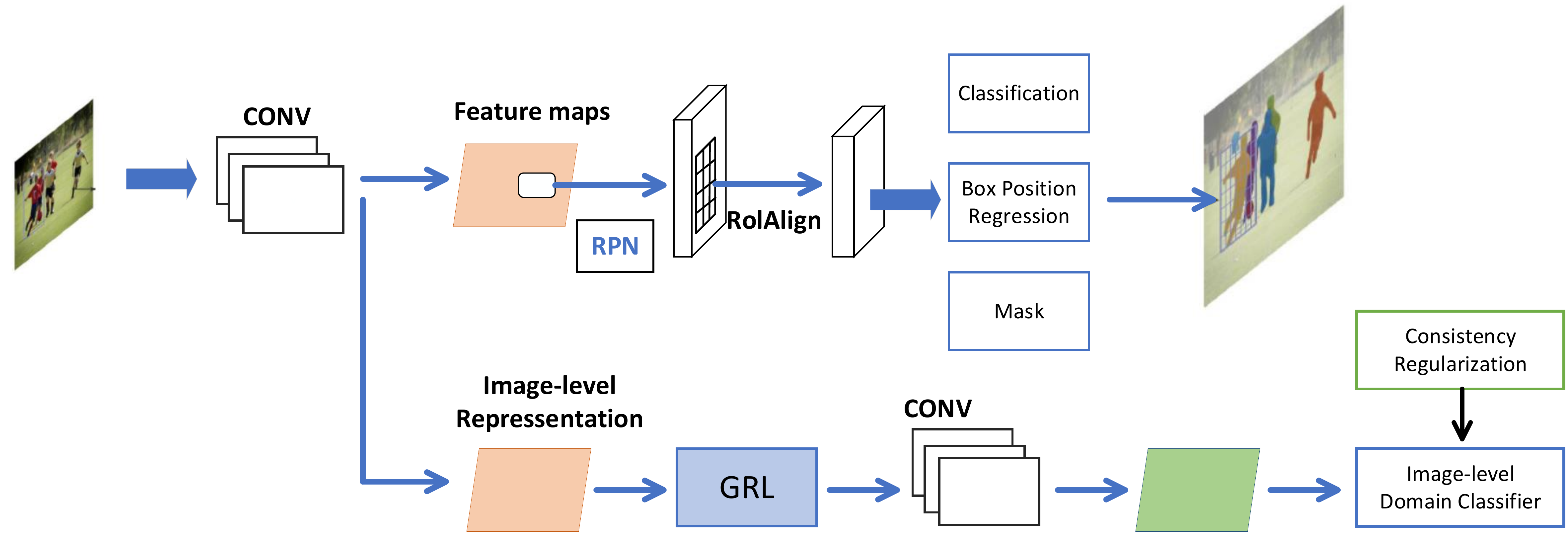}
\caption{DMask-RCNN structure.}
\label{fig:DMask}
\end{figure}



\begin{equation}
-\sum_{i} (y_{i}log(p_i)+(1-y_i)log(1-p_i)),
\end{equation}
\noindent where $y_i$ is the domain label of the $i_{th}$ image, and $p_i$ is the prediction probability from the domain classifier. The overall loss of DMask-RCNN can therefore be written as:

\begin{equation}
\label{total loss}
\begin{split}
L(\theta_{res}, \theta_{head}, \theta_{domain}) & = L_{C,B}(C,B| \theta_{res}, \theta_{head}, x \in D_s )\\
& - \lambda L_d(G_d| \theta_{res}, x\in D_s, D_t) \\
& + \lambda L_d(G_d| \theta_{domain}, x\in D_s, D_t),
\end{split}
\end{equation}

\noindent where $x$ is the input image, and $D_s$ and $D_t$ represents the source and target domain, respectively. $\theta$ denotes the corresponding weights of each network component. $G$ represents the mapping function of the feature extractor; $I$ is the feature map distribution; $B$ is the bounding box of an object and $C $ is the object class. Note that when calculating the $L_{C, B}$, only source domain inputs will be counted in since the target domain has no labels. 

As seen from Eqn. (\ref{total loss}), the negative gradient of the domain classifier loss needs to be propagated back to ResNet, whose implementation relies on the gradient reverse layer~\cite{ganin2014unsupervised} (GRL, Fig.~\ref{fig:DMask}). The GRL is added after the feature maps generated by the ResNet and feeds its output to the domain classifier. This GRL has no parameters except for the hyper-parameter $\lambda$, which, during forward propagation, acts as an identity transform. However, during back propagation, it takes the gradient from the upper level and multiplies it by $-\lambda$ before passing it to the preceding layers. 






\paragraph{Experiments} To train DMask-RCNN, MS COCO (clean images) were always used as the source domain, while \textbf{two target domain options} were designed to consider two types of domain gap: (1) all unannotated realistic haze images from RESIDE; and (2) dehazed results of those unannotated  images, using MSCNN \cite{ren2016single}. The corresponding DMask-RCNNs are called DMask-RCNN1 and DMask-RCNN2, respectively. 

We initialized the Mask-RCNN component of DMask-RCNN with a pre-trained model on MS COCO. All models were trained for 50, 000 iterations with learning rate 0.001, then another 20, 000 iterations with learning rate 0.0001. We used a naive batch size of 2, including one image randomly selected from the source domain and the other from the target domain, noting that larger batches may further benefit performance. We also tried to concatenate dehazing pre-processing (AOD-Net and MSCNN) with DMask-RCNN models to form new dehazing-detection cascades.


\begin{table}[t]
\begin{center}
\begin{tabular}{|l|c|}
\hline
\textbf{Pipelines} & \textbf{mAP} \\ 
 \hline
DMask-RCNN1 & 0.612 \\ \hline
DMask-RCNN2 & 0.617 \\ \hline
\hline
AOD-Net + DMask-RCNN1 & 0.602 \\ \hline
AOD-Net + DMask-RCNN2 & 0.605 \\ \hline
\hline
MSCNN + Mask-RCNN & \textcolor{blue}{0.626} \\ \hline
MSCNN + DMask-RCNN1 & \textcolor{green}{0.627} \\ \hline
MSCNN + DMask-RCNN2 & \textcolor{red}{0.634} \\ \hline
\end{tabular}
\end{center}
\caption{Solution set 2 mAP results on RTTS. Top 3 results are colored in red, green, and blue, respectively.}
\label{detection2}
\end{table}

Table \ref{detection2} shows the results of solution set 2 (the naming convention is the same as in Table \ref{detection1}), from which we can conclude that:

\begin{itemize}
\item the domain-adaptive detector presents a very promising approach, and its performance significantly outperforms the best results in Table \ref{detection1};\footnote{By saying that, we also emphasize that Table \ref{detection1} results have not undergone joint tuning as in \cite{liu2017enhance,li2017aod}, so there is potential for further improvements.} 

\item the power of strong detection models (Mask-RCNN) is fully exploited, given the proper domain adaptation, in contrast to the poor performance of vanilla Mask RCNN in Table \ref{detection1};

\item DMask-RCNN2 is always superior to DMask-RCNN1, showing that the choice of dehazed images as the target domain matters. We make the reasonable hypothesis that the domain discrepancy between dehazed and clean images is smaller than that between hazy and clean images, so DMask-RCNN performs better when the existing domain gap is narrower; and

\item the best result in solution set 2 is from a dehazing + detection cascade, with MSCNN as the dehazing module and DMask-RCNN as the detection module and highlighting: \textbf{the joint value of dehazing pre-processing and domain adaption}.
\end{itemize}

\section{Conclusion}
This paper tackles the challenge of single image dehazing and its extension to object detection in haze. The solutions are proposed from diverse perspectives ranging from novel loss functions (Task 1) to enhanced dehazing-detection cascades as well as domain-adaptive detectors (Task 2). By way of careful experiments, we significantly improve the performance of both tasks, as verified on the RESIDE dataset. We expect further improvements as we continue to study this important dataset and tasks.


\section*{Acknowledgements}
The study was initially performed as a team project effort in the Machine Learning course (\href{http://people.tamu.edu/~atlaswang/18CSCE633.html}{Spring 2018, CSCE 633}) of CSE@TAMU, taught by Dr. Zhangyang Wang. We acknowledge Texas A\&M High Performance Research Computing (HPRC) for providing some of the computing resources used in this research.

\bibliographystyle{ieeetr}
\bibliography{main}

\end{document}